\documentclass[twocolumn]{article}
\usepackage[a4paper, left=1.5cm, right=1.5cm, top=2cm, bottom=2cm]{geometry}
\usepackage{graphicx}
\usepackage{booktabs}
\usepackage{hyperref}
\usepackage{tabularx}
\usepackage{array}
\usepackage{cite}
\usepackage{tikz} 
\usepackage{amsmath}
\usepackage{xcolor}
\usetikzlibrary{positioning, arrows, shapes.geometric}
\usepackage{listings}
\usepackage{xcolor}
\usepackage{braket}
\definecolor{lightgray}{gray}{0.95}
\newcommand{\keywords}[1]{\textbf{Keywords:} #1}

\usepackage{algorithm}
\usepackage{algpseudocode}
\usepackage{float}    
\usepackage{ragged2e} 

\begin{document}

\title{Quantum Feature Optimization for Enhanced Clustering of Blockchain Transaction Data \thanks{The views expressed in this article are those of the authors and do not represent the views of Wells Fargo. This article is for informational purposes only. Nothing contained in this article should be construed as investment advice. Wells Fargo makes no express or implied warranties and expressly disclaims all legal, tax, and accounting implications related to this article. }}

\author{
Yun-Cheng Tsai$^1$, Samuel Yen-Chi Chen$^2$\\
$^1$National Taiwan Normal University $\quad$ $^2$Wells Fargo\\
\small \texttt{pecu@ntnu.edu.tw, ycchen1989@ieee.org}
}

\maketitle

\begin{abstract}
Blockchain transaction data exhibits high dimensionality, noise, and intricate feature entanglement, presenting significant challenges for traditional clustering algorithms. In this study, we conduct a comparative analysis of three clustering approaches: (1) Classical K-Means Clustering, applied to pre-processed feature representations; (2) Hybrid Clustering, wherein classical features are enhanced with quantum random features extracted using randomly initialized quantum neural networks (QNNs); and (3) Fully Quantum Clustering, where a QNN is trained in a self-supervised manner leveraging a SwAV-based loss function to optimize the feature space for clustering directly. The proposed experimental framework systematically investigates the impact of quantum circuit depth and the number of learned prototypes, demonstrating that even shallow quantum circuits can effectively extract meaningful non-linear representations, significantly improving clustering performance.
\end{abstract}


\section{Introduction}
Blockchain transaction data is complex and high-dimensional, posing significant challenges for traditional clustering methods~\cite{Xu2005,Li2020,Khan2021}. Historically, clustering techniques applied to blockchain transactions have predominantly relied on classical machine learning methods, such as K-Means and DBSCAN. These conventional techniques, however, struggle to handle the high-dimensional, sparse, and entangled nature of blockchain datasets. The reliance on Euclidean distance metrics further limits their ability to capture the complex, non-linear relationships necessary for detecting anomalies and fraudulent activities~\cite{Liu2019,Wittek2014}.

Quantum machine learning (QML) introduces novel capabilities to address these challenges. Quantum neural networks (QNNs) and variational quantum algorithms (VQA) leverage uniquely quantum phenomena such as superposition and entanglement to perform non-linear transformations, remapping data into higher-dimensional Hilbert spaces where latent structures become more discernible~\cite{Biamonte2017,Schuld2015,Aharon2020,Cerezo2021}. These quantum feature maps enable the extraction of correlations that classical techniques often overlook, making them particularly promising for blockchain analytics~\cite{Havlicek2019,Arrazola2020}. Several studies have emphasized the potential of quantum algorithms to tackle challenges associated with high-dimensional and noisy datasets~\cite{Dunjko2018,Preskill2018,Schuld2020}.

The convergence of QML and blockchain analytics presents a fertile ground for innovation. By leveraging quantum feature optimization, our work aims to advance clustering methodologies and develop more robust analytical tools tailored to the complexities of blockchain data~\cite{Mitarai2018,Benedetti2019,Barkoutsos2020,Wang2021}. Despite these promising developments, the literature still lacks a systematic comparison among purely classical, hybrid quantum-classical, and fully quantum clustering approaches. This omission is particularly significant given the challenges posed by high dimensionality, noise, and feature entanglement in blockchain datasets.

To bridge this gap, we conduct a comprehensive comparative analysis of three clustering strategies:
\begin{enumerate}
    \item \textbf{Traditional K-Means:} Clustering applied directly to pre-processed classical features.
    \item \textbf{Hybrid Quantum-Classical Approach:} Augmenting classical features with additional quantum features extracted via an untrained (random) quantum neural network.
    \item \textbf{Fully Quantum Clustering:} Employing a quantum neural network trained end-to-end with a composite loss function to optimize the feature space for clustering.
\end{enumerate}

Furthermore, we integrate a SwAV-based self-supervised learning loss function within a QNN framework to enhance quantum feature optimization for clustering blockchain data~\cite{Caron2020}. Our results demonstrate that even shallow quantum circuits (quantum depth = 1) yield significant improvements in clustering performance, underscoring the potential of quantum-assisted techniques in handling the intricate nature of blockchain transaction data. This study highlights the advantages of QML in blockchain analytics and lays the groundwork for future advancements in robust, scalable, and interpretable clustering methodologies.

Recent works under the ``Quantum-Train'' framework \cite{liu2024quantum,lin2024quantum,liu2024qtrl,liu2024federated,liu2024quantum2,lin2024quantum2,liu2024quantum3,chen2024_QT_DIST_RL,liu2024_QT_FWP} propose using untrained QNNs to generate weights for classical neural networks, enabling model compression and training acceleration. Our work extends this line by exploring how quantum features themselves, both random and optimized, can directly enhance unsupervised clustering—thereby shifting from parameter compression to feature representation learning.

\section{Methodology}
\subsection{Dataset Description}
We use blockchain transaction data containing block numbers, sender/recipient addresses, and token information. Categorical attributes are label-encoded, and numerical attributes are standardized via RobustScaler.

\subsection{Clustering Approaches}
\begin{enumerate}
\item \textbf{Traditional K-Means:} Clustering is performed directly on pre-processed classical features.
\item \textbf{Quantum Random Feature Extraction:} An untrained quantum neural network generates additional quantum features concatenated with classical features for K-Means clustering.
\item \textbf{Fully Quantum Clustering:} A quantum neural network is trained using a SwAV loss function to optimize clustering quality directly.
\end{enumerate}

\begin{figure*}[ht]
\centering
\begin{tikzpicture}[node distance=1.8cm and 2.5cm, auto] 

    \tikzstyle{startstop} = [rectangle, rounded corners, minimum width=3cm, minimum height=1cm, text centered, draw=black, fill=red!30]
    \tikzstyle{process} = [rectangle, minimum width=3.5cm, minimum height=1cm, text centered, draw=black, fill=orange!30]
    \tikzstyle{arrow} = [thick,->,>=stealth]

    \node (start) [startstop] {Start};
    \node (preproc) [process, below of=start] {Data Collection};

    \node (trad) [process, below left of=preproc, xshift=-3.5cm, yshift=-0.6cm] {Classical Features};
    \node (qrand) [process, below of=preproc] {Quantum Random Feature Extraction};
    \node (qtrain) [process, below right of=preproc, xshift=4cm, yshift=-0.6cm] {Train QNN (Composite Loss)};

    \node (kmeans1) [process, below of=trad, yshift=-0.5cm] {K-Means};
    \node (comb2) [process, below of=qrand] {Hybrid Feature Set};
    \node (qclust) [process, below of=qtrain, yshift=-0.5cm, xshift=0cm] {Quantum Clustering}; 

    \node (kmeans2) [process, below of=comb2] {K-Means};

    \node (eval) [process, below of=kmeans2, yshift=0cm] {Evaluation}; 
    \node (result) [startstop, below of=eval] {Results \& Visualization};

    \draw [arrow] (start) -- (preproc);
    \draw [arrow] (preproc) -- (trad);
    \draw [arrow] (preproc) -- (qrand);
    \draw [arrow] (preproc) -- (qtrain);

    \draw [arrow] (trad) -- (kmeans1);
    \draw [arrow] (kmeans1.south) -- ++(0,-0.5) |- (eval.west); 

    \draw [arrow] (qrand) -- (comb2);
    \draw [arrow] (comb2) -- (kmeans2);
    \draw [arrow] (kmeans2.south) -- ++(0,-0.5) -- (eval.north); 

    \draw [arrow] (qtrain) -- (qclust);
    \draw [arrow] (qclust.south) -- ++(0,-0.5) |- (eval.east); 

    \draw [arrow] (eval) -- (result);

\end{tikzpicture}
\caption{Flowchart of the experimental design comparing three approaches: (1) Traditional K-Means, (2) Quantum Random Feature Extraction + K-Means, and (3) Fully Quantum Clustering.}
\label{fig:flowchart}
\end{figure*}
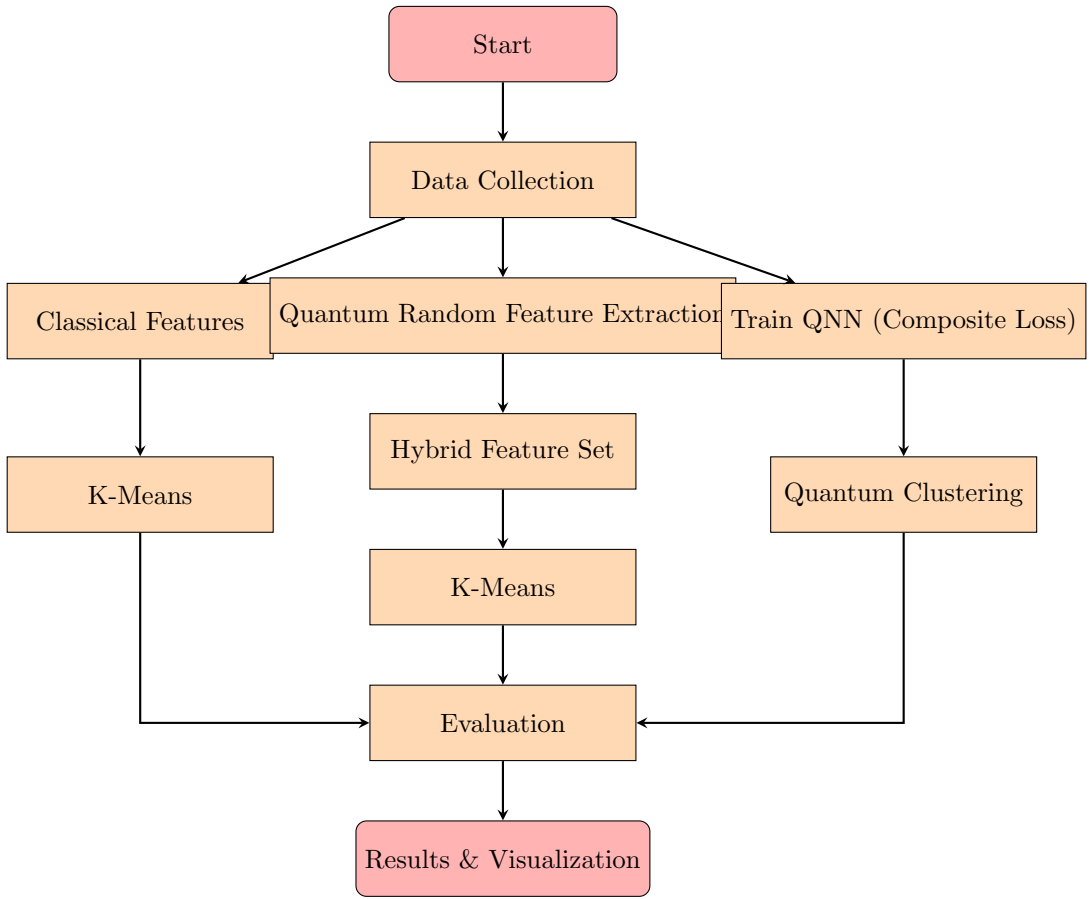

\noindent \textbf{Explanation of Figure~\ref{fig:flowchart}:} This flowchart outlines the overall experimental design. It details the process from data collection to feature processing, which branches into three paths: classical feature extraction, quantum random feature extraction, and QNN training using a SwAV loss. Each branch leads to clustering and subsequent evaluation, highlighting how integrating quantum and classical methods contributes to enhanced clustering performance.

Our experimental framework comprises four main stages: Data collection, Quantum model processing, clustering, and evaluation. Figure~\ref{fig:flowchart} presents a flow diagram of the overall experimental design.

\subsection{Detailed Method Description}
\subsubsection{Data Preprocessing}
Raw blockchain transaction data is first cleaned to remove incomplete or erroneous records. The key features are then extracted and processed:
\begin{itemize}
    \item \textbf{Feature Selection:} Extract attributes such as \textit{BlockNumber}, \textit{TimeStamp}, \textit{Hash}, \textit{From}, \textit{To}, \textit{Value}, \textit{TokenName}, and \textit{TokenSymbol}.
    \item \textbf{Encoding and Normalization:} Categorical features (e.g., addresses, token names) are encoded using Label Encoding, while numerical features are standardized using robust scaling.
\end{itemize}

\subsubsection{Approach 1: Traditional K-Means}
The classical approach involves applying K-Means clustering directly to the pre-processed features. The objective function minimized is the standard within-cluster sum of squares (WCSS):
\begin{equation}
L_{\text{WCSS}} = \sum_{i=1}^{k}\sum_{x \in C_i} \|x - \mu_i\|^2,
\end{equation}
where \(C_i\) denotes the \(i\)th cluster and \(\mu_i\) its centroid.

\subsubsection{Approach 2: Quantum Random Feature Extraction + K-Means}

A QNN with $N$ qubits can generate up to $2^N$ numerical values when measured, capturing the probability distribution over all computational basis states $\ket{00 \cdots 00}$ to $\ket{11 \cdots 11}$. For quantum feature extraction, we employ a variational quantum circuit $V(\Theta)$, where each layer consists of parameterized rotation gates $R_y(\theta)$ applied to all qubits, followed by CNOT gates implementing. The parameters $V(\Theta)$ are randomly initialized and not updated during inference. The output distribution is used to derive quantum features concatenated with classical features for downstream clustering.

\begin{figure}[htbp]
\centering
\includegraphics[width=1\columnwidth]{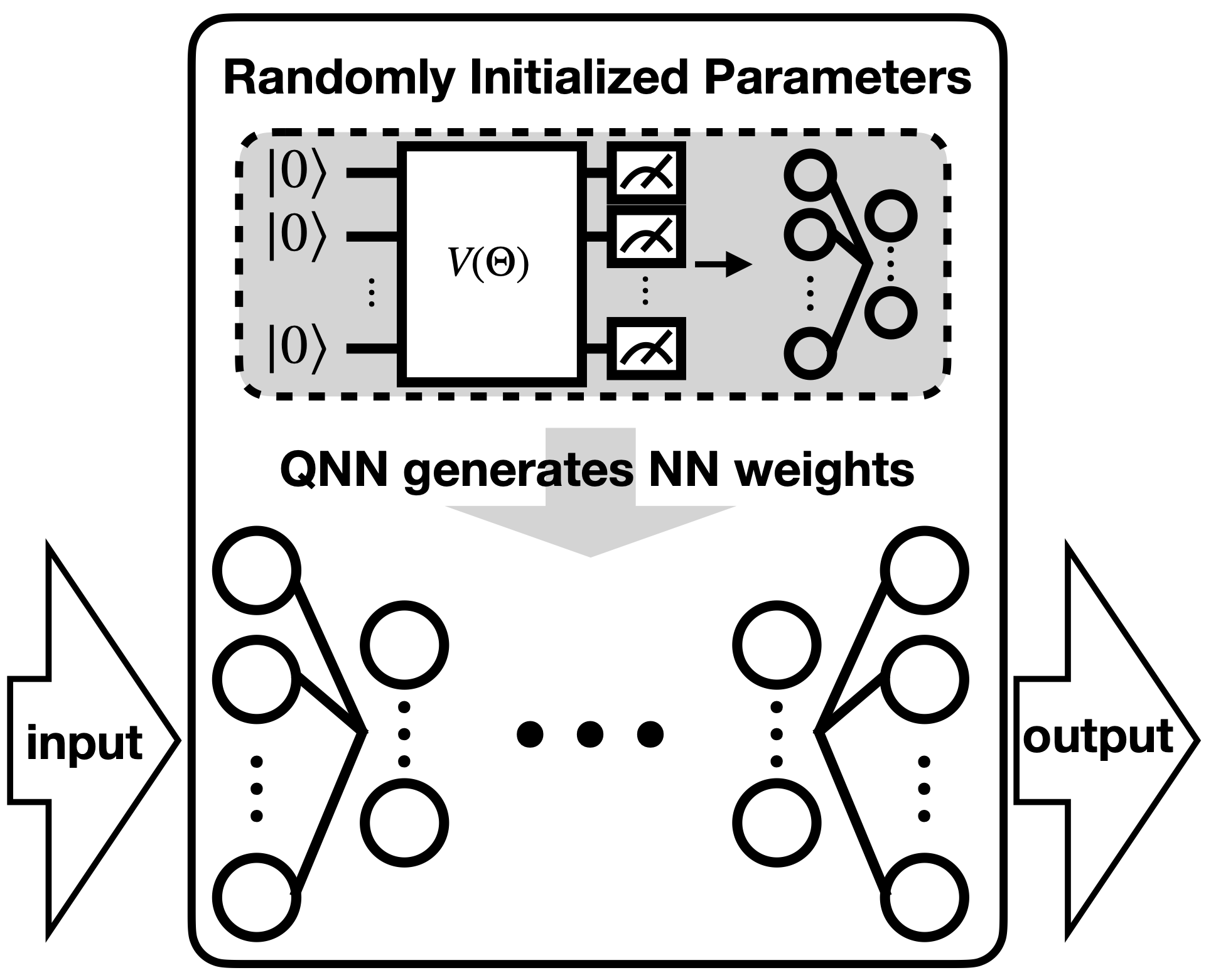}
\caption{Random QNN for generating weights for feature transformation NN. This figure depicts a randomly initialized variational quantum circuit $V(\Theta)$, composed of single-qubit $R_y$ rotations and CNOT entanglement. The output distribution is used as quantum features.}
\label{fig:random_QNN}
\end{figure}

Prior research has demonstrated that this property of QNNs can be leveraged to generate weights for classical neural networks (NNs), enabling efficient parameter compression \cite{liu2024quantum,lin2024quantum,liu2024qtrl,liu2024federated,liu2024quantum2,lin2024quantum2,liu2024quantum3,chen2024_QT_DIST_RL,liu2024_QT_FWP}. In this work, we explore the capability of an untrained (randomly initialized) QNN to generate weights for a classical NN, which serves as a feature extractor in a clustering task. We refer to these extracted features as \emph{quantum features}. The quantum features are normalized and concatenated with classical features to construct a hybrid feature representation. K-means clustering is then applied to this enriched feature space. The proposed scheme is illustrated in Figure~\ref{fig:random_QNN}.
\subsubsection{Approach 3: Fully Quantum Clustering}
\begin{figure}[htbp]
\centering
\includegraphics[width=1\columnwidth]{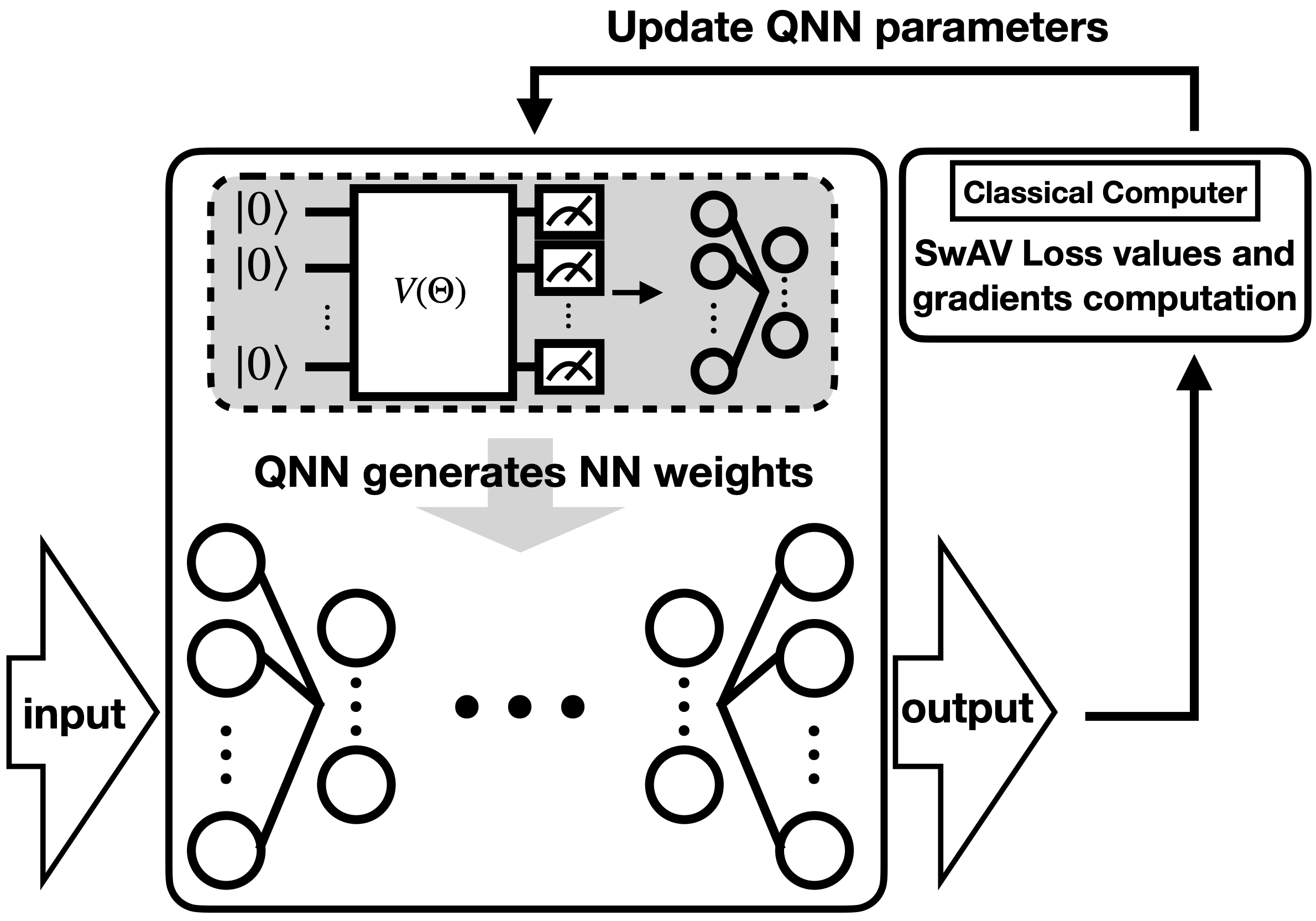}
\caption{Trained QNN for generating weights for feature transformation NN. This figure illustrates a trained quantum circuit $V(\Theta)$, whose parameters are optimized via SwAV loss. The circuit uses the same gate structure as in Figure~\ref{fig:random_QNN} but undergoes end-to-end training to enhance clustering performance.}
\label{fig:trained_QNN}
\end{figure}

In this approach, we further investigate whether optimizing the QNN parameters can enhance clustering performance. Under the hybrid quantum-classical paradigm, the QNN is integrated with a SwAV loss module, forming a unified model. The entire hybrid model is trained end-to-end, utilizing standard backpropagation for optimization. This scheme is illustrated in Figure~\ref{fig:trained_QNN}.

This approach's variational quantum circuit $V(\Theta)$ adopts a layered ansatz with trainable parameters. Each layer contains single-qubit rotations $R_y(\theta)$ and CNOT-based entangling adjacent qubits. During training, the parameters $\Theta$ are optimized via backpropagation through the SwAV loss function.

The SwAV loss is defined as:
\begin{equation}
L_{\text{SwAV}} = \text{KL}\left(\text{log}(\text{Softmax}(Z_1/\tau)) \,\|\, P_2\right),
\end{equation}
where $Z_1$ denotes logits from one augmentation, temperature $\tau$ controls output sharpness. The target distribution $P_2$ is obtained by smoothing the prototype assignment $c_k$ into a soft label:
\begin{equation}
P_2(i) = (1-\epsilon)\cdot\delta_{i,k}+\frac{\epsilon}{N_p},
\end{equation}
with label smoothing factor $\epsilon=0.1$, $N_p$ being number of prototypes, and $\delta$ the Kronecker delta.

This loss guides the QNN in learning robust feature representations that are subsequently used for K-means clustering. Training is performed using the Adam optimizer with gradient clipping to mitigate gradient explosion. The Algorithm~\ref{alg:qnn_swav} outlines the fully quantum clustering approach.

\begin{algorithm}[htbp]
\caption{Quantum Neural Network Training with SwAV Loss}
\label{alg:qnn_swav}
\raggedright
\begin{algorithmic}[1]
\For{each $num\_prototypes$ in $prototype\_range$}
    \State Create output directories for current $num\_prototypes$
    \For{each $quantum\_depth$ in $quantum\_depth\_range$}
        \State Initialize base neural network model
        \State Initialize QNN with given $quantum\_depth$ using the base model
        \State Initialize the SwAV loss module with current $num\_prototypes$
        \State Set optimizer with parameters from QNN and SwAV loss

        \For{each epoch in $num\_epochs$}
            \For{each batch in $scaled\_features$}
                \State Set QNN to training mode
                \State $quantum\_output \gets QNN(batch)$
                \State Compute $P_1\gets Softmax(aug_1 / \tau)$ with $\tau = 0.07$
                \State Compute $P_2\gets$ Smoothed one-hot label for prototype assignment $(\epsilon=0.1$
                \State $loss \gets \text{KL}(P_1 || P_2)$
                \State optimizer.zero\_grad()
                \State Backpropagate $loss$
                \State Apply gradient clipping
                \State optimizer.step()
            \EndFor

            \State Set QNN to evaluation mode
            \State Initialize empty list $quantum\_features$
            \For{each batch in $scaled\_features$}
                \State $quantum\_output \gets QNN(batch)$ without gradient computation
                \State Append $quantum\_output$ to $quantum\_features$
            \EndFor
            \State Concatenate $quantum\_features$
        \EndFor
    \EndFor
\EndFor
\end{algorithmic}
\end{algorithm}

\section{Experiments and Results}
\subsection{Performance Metrics}
The clustering performance is evaluated using:
\begin{itemize}
\item \textbf{Silhouette Score} - Measures how well-separated clusters are.
\item \textbf{Davies-Bouldin Index} - A lower value indicates better clustering quality.
\item \textbf{Calinski-Harabasz Index} - A higher value indicates well-defined clusters.
\end{itemize}

Table~\ref{tab:best_results} presents the best clustering performance for different cluster numbers ($K=2$ to $K=6$), comparing Monte Carlo at Depth=0 and Depth=1 with QNN.

\begin{table*}[ht]
\centering
\caption{Comparison of Best Clustering Performance: Monte Carlo (Quantum Features) vs QNN for $K=2$ to $K=6$}
\begin{tabular}{ccccccc}
\toprule
$K$ & Method & Depth & Epoch & Silhouette Score & Davies-Bouldin & Calinski-Harabasz \\
\midrule
2 & Classical Features & - & - & 0.265773 & 1.525312 & 1,605 \\
3 & Quantum Features (Worst Run) & 1 & -  & 0.394566 & 0.927151 & 3,847 \\
2 & Quantum Features (Average) & 1 & -  & 0.660680 & 0.470618 & 133,976 \\
2 & Quantum Features (Best Run) & 1 & -  & 0.996368 & 0.042230 & 1,260,727 \\
2 & QNN              & 2 & 1  & \textbf{0.999777}        & \textbf{1.111477e-8}        & \textbf{15,833,657,123,341}         \\
\midrule
3 & Classical Features & - & - & 0.250360 & 1.473543 & 1,347 \\
3 & Quantum Features (Worst Run) & 1 & -  & 0.383393 & 0.868310 & 4,066 \\
3 & Quantum Features (Average) & 1 & -  & 0.612198 & 0.553668 & 16,307,927,154 \\
3 & Quantum Features (Best Run) & 1 & -  & \textbf{0.999994} & \textbf{0.000798} & 456,616,095,717 \\
3 & QNN                          & 4 & 5 & 0.999510        & 0.141294       & \textbf{168,665,013,271,780}           \\
\midrule
4 & Classical Features & - & - & 0.242149 & 1.413766 & 1,215 \\
4 & Quantum Features (Worst Run) & 1 & -  & 0.351097 & 0.935585 & 3,891 \\
4 & Quantum Features (Average) & 1 & -  & 0.600428 & 0.577474 & 45,602,215,148 \\
4 & Quantum Features (Best Run) & 1 & -  & 0.998005 & 0.223911 & 1,276,852,339,920 \\
4 & QNN                          & 9 & 9  & \textbf{0.999246}       & \textbf{0.000054}        & \textbf{1,307,035,577,888,410}           \\
\midrule
5 & Classical Features & - & - & 0.240189 & 1.433870 & 1,148 \\
5 & Quantum Features (Worst Run) & 1 & -  & 0.315429 & 1.058288 & 3,711 \\
5 & Quantum Features (Average) & 1 & -  & 0.600791 & 0.567613 & 61,140,087,483 \\
5 & Quantum Features (Best Run) & 1 & -  & 0.997542 & 0.305401 & 1,711,909,181,014 \\
5 & QNN                          & 6 & 1  & \textbf{0.998789}        & \textbf{0.138369}        & \textbf{7,704,660,919,690,170}           \\
\midrule
6 & Classical Features & - & - & 0.240189 & 1.433870 & 1,149 \\
6 & Quantum Features (Worst Run) & 1 & -  & 0.328886 & 0.993762 & 3,561 \\
6 & Quantum Features (Average) & 1 & -  & 0.597184 & 0.580199 & 84,564,962,473 \\
6 & Quantum Features (Best Run) & 1 & -  & 0.997905 & 0.262240 & 2,367,802,555,468 \\
6 & QNN                          & 6 & 1  & \textbf{0.998540}        & \textbf{0.000125}        & \textbf{23,337,479,575,752,000}           \\
\bottomrule
\end{tabular}
\label{tab:best_results}
\end{table*}

\noindent \textbf{Explanation of Table~\ref{tab:best_results}:} In this table, we compare the clustering performance of various approaches across different cluster numbers ($K=2$ to $K=6$). The column \textit{Method} indicates the technique used (e.g., Classical Features, Quantum Features, QNN). The \textit{Depth} column refers to the quantum circuit depth, while \textit{Epoch} represents the number of training iterations for the QNN. The performance metrics are defined as follows:
\begin{itemize}
    \item \textbf{Silhouette Score:} Higher values denote better cluster separation.
    \item \textbf{Davies-Bouldin Index:} Lower values indicate more compact clusters.
    \item \textbf{Calinski-Harabasz Index:} Higher values reflect well-separated clusters.
\end{itemize}
The table demonstrates that even with a shallow quantum circuit (Depth = 1), quantum feature extraction significantly enhances clustering performance compared to classical features, and that further training with QNN provides additional improvements.

The results indicate that different quantum depths lead to varying optimal cluster counts. Notably, deeper quantum models achieve significantly higher Silhouette Scores and lower Davies-Bouldin values, confirming their effectiveness in feature optimization.

\begin{figure}[!htbp]
\centering
\includegraphics[width=0.5\textwidth]{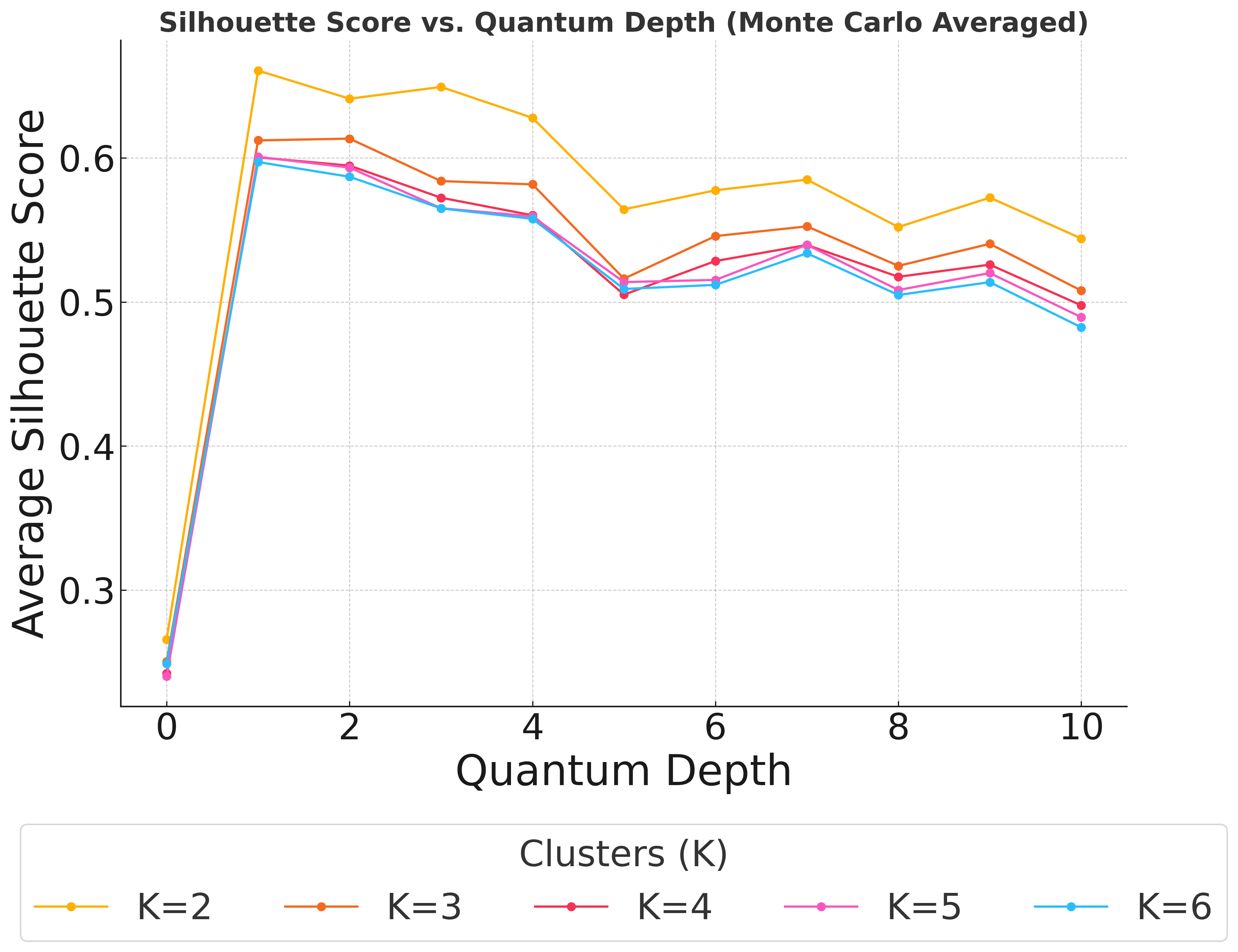}
\caption{Silhouette Score vs. Quantum Depth}
\label{fig:silhouette}
\end{figure}

\begin{figure}[!htbp]
\centering
\includegraphics[width=0.5\textwidth]{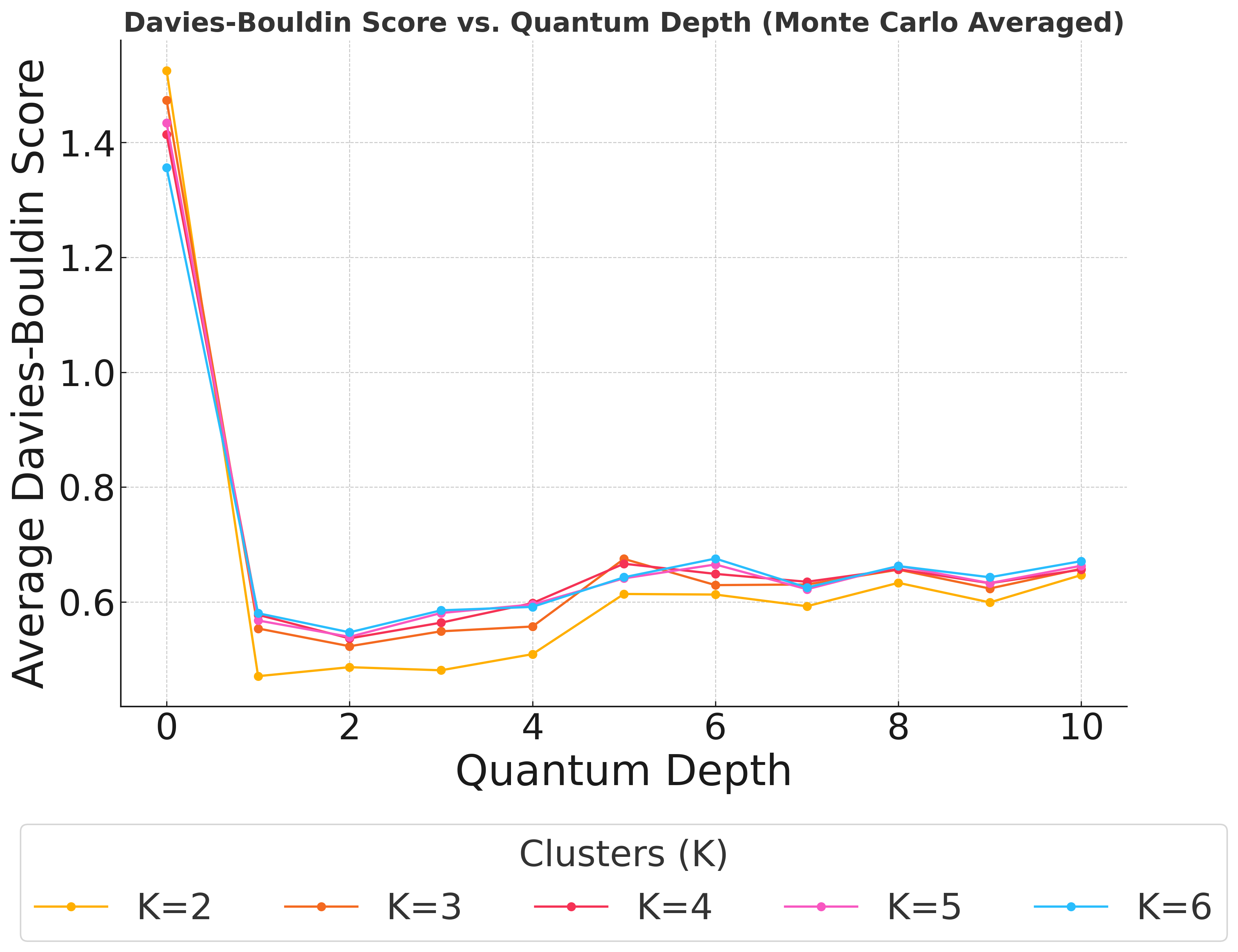}
\caption{Davies-Bouldin Index vs. Quantum Depth}
\label{fig:daviesbouldin}
\end{figure}

\begin{figure}[!htbp]
\centering
\includegraphics[width=0.5\textwidth]{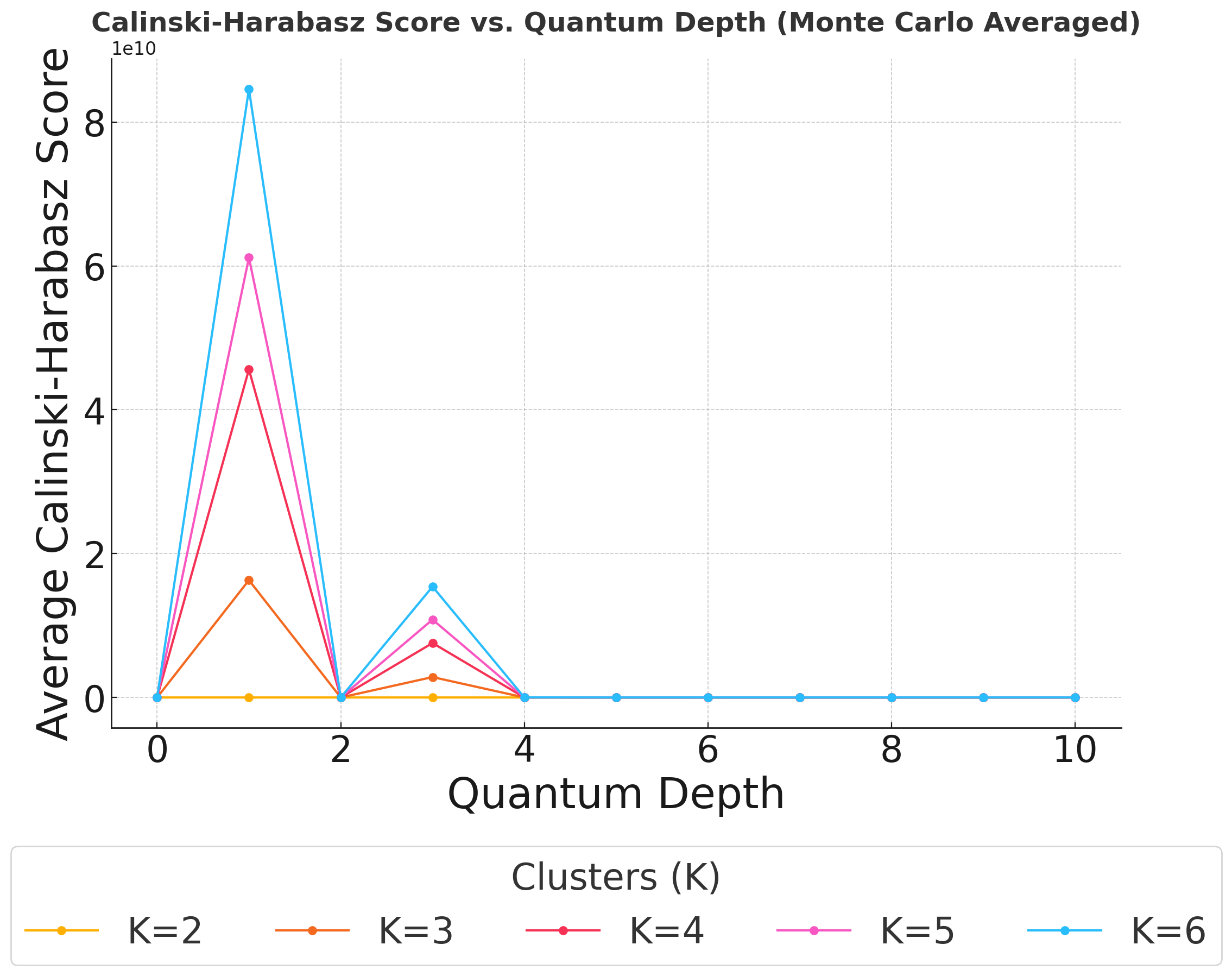}
\caption{Calinski-Harabasz Index vs. Quantum Depth}
\label{fig:calinski}
\end{figure}

Figures~\ref{fig:silhouette}, ~\ref{fig:daviesbouldin}, and ~\ref{fig:calinski} illustrate how clustering performance varies with quantum depth. The results indicate that quantum-enhanced features significantly improve clustering outcomes even at lower depths.

The following factors can explain the superior performance at quantum depth 1:
\begin{itemize}
    \item \textbf{Non-linear Mapping:} Even a shallow quantum circuit implements non-linear transformations that map raw blockchain features into a higher-dimensional Hilbert space. This transformation exposes subtle correlations and latent patterns not captured in the original space.
    \item \textbf{Mitigation of Noise and Entanglement:} Blockchain data often suffers from noisy and entangled features due to its inherent structure. Quantum feature extraction effectively “untangles” these features, leading to clearer cluster separability.
    \item \textbf{Feature Space Expansion:} By expanding the feature space, the quantum model alleviates the curse of dimensionality and enables the clustering algorithm to better discriminate between different transaction patterns.
\end{itemize}
In contrast, traditional clustering on raw features fails to capture these complex relationships, resulting in poorer clustering performance.

\noindent \textbf{Explanation of Figure~\ref{fig:silhouette}:} This figure displays the variation of the Silhouette Score as a function of quantum circuit depth. Higher Silhouette Scores correspond to better-defined clusters. The trend observed here indicates that even a shallow quantum circuit (quantum depth = 1) can significantly improve cluster separation, supporting the effectiveness of quantum feature mapping.

\noindent \textbf{Explanation of Figure~\ref{fig:daviesbouldin}:} This figure illustrates the Davies-Bouldin Index across different quantum depths. Lower values of the Davies-Bouldin Index indicate better clustering quality. The results confirm that quantum-enhanced features lead to more compact clusters, thereby reducing the Davies-Bouldin Index.

\noindent \textbf{Explanation of Figure~\ref{fig:calinski}:} This figure presents the Calinski-Harabasz Index as a function of quantum circuit depth. Higher Calinski-Harabasz values are indicative of well-separated and distinct clusters. The upward trend observed with increasing quantum depth reinforces the positive impact of quantum feature optimization on cluster differentiation.

\section{Discussion}
Our experimental results show that incorporating quantum feature optimization into clustering frameworks significantly enhances performance on complex blockchain datasets. Both approaches—extracting quantum features using a shallow circuit (quantum depth = 1) without any training and training a quantum neural network (QNN) with deeper circuits—consistently outperform the classical baseline across key metrics (Silhouette Score, Davies-Bouldin Index and Calinski-Harabasz Index). For instance, while the QNN, after multi-layer training, achieves exceptionally high scores (e.g., a Silhouette Score of 0.999777 for K=2K=2), the mere extraction of quantum features via a shallow circuit already yields results that are remarkably close to those of the trained QNN. Even the worst-case quantum feature extraction consistently surpasses the performance of conventional methods.

This indicates that by projecting classical data into a higher-dimensional Hilbert space, the quantum feature mapping effectively uncovers latent correlations and patterns obscured in the original high-dimensional space. In this expanded feature space, subtle non-linear relationships become more discernible, thus facilitating better cluster formation. While QNN training further refines this process by optimizing the model parameters, it also introduces additional computational overhead. Therefore, in applications where computational efficiency is crucial, extracting quantum features can be a highly effective alternative, achieving near-optimal clustering performance with minimal training.

\section{Conclusion}
In conclusion, our study demonstrates the transformative potential of quantum feature optimization for clustering complex blockchain transaction data. By comparing traditional K-Means clustering with a hybrid quantum-classical approach (which solely extracts quantum features) and a fully quantum approach (using QNN training), we show that even a shallow quantum circuit can improve clustering performance. The QNN, with its additional training, further elevates performance by leveraging non-linear transformations to expand the feature space and reveal hidden data patterns. However, our findings also highlight that extracting quantum features without extra training for many scenarios provides nearly comparable results, thus offering a resource-efficient solution. These insights pave the way for future research into optimizing quantum circuit architectures and developing adaptive hybrid models tailored to large-scale blockchain data's computational demands and complexities.



\end{document}